# TRUTH VALIDATION WITH EVIDENCE

Papis Wongchaisuwat and Diego Klabjan

Department of Industrial Engineering and Management Sciences, Northwestern University, Evanston, IL

## Abstract

In the modern era, abundant information is easily accessible from various sources, however only a few of these sources are reliable as they mostly contain unverified contents. We develop a system to validate the truthfulness of a given statement together with underlying evidence. The proposed system provides supporting evidence when the statement is tagged as false. Our work relies on an inference method on a knowledge graph (KG) to identify the truthfulness of statements. In order to extract the evidence of falseness, the proposed algorithm takes into account combined knowledge from KG and ontologies. The system shows very good results as it provides valid and concise evidence. The quality of KG plays a role in the performance of the inference method which explicitly affects the performance of our evidence-extracting algorithm.

## 1.INTRODUCTION

Accessing information online is expanding tremendously in various domains as reported in a study by the Pew Internet Project's research (Anderson and Perrin, 2016). According to the study, offline population in the U.S. has declined significantly since 2000 to the extent that in 2016 only 13% of U.S. adults did not use the internet. Internet usage gives people an opportunity to extensively seek information online; however, posted contents available on web pages are not necessarily reliable. As online information spreads rapidly, its quality is considerably crucial. Misinformation potentially leads to serious consequences significantly affecting internet users. The main motivation of our study is to validate the truthfulness of textual information obtained from various sources as well as to provide supporting evidence.

Humans can identify the truthfulness of a statement particularly for common fact cases. Nevertheless, manually inspecting statements is a time-consuming process that becomes impossible for large-scale data. Determining the truthfulness of each statement in an automated fashion is a promising alternative solution. This problem is highly challenging due to the lack of an encompassing and comprehensive corpora of all true statements. Despite of its challenges, it draws a lot of attention from prior studies to develop truthfulness-validating systems. The previously proposed systems mainly rely on web search engines to verify whether statements are true or false. Additional information regarding sources which statements are extracted from is also taken into account in most algorithms.

In comparison to these truthfulness-validating systems, our work relies on knowledge from reliable sources rather than web search engines. Statements gathered from reliable sources have various length and may be verbose and thus we represent each of these statements as triplets consisting of a subject entity, an object entity, and their relation. These triplets capture the main contents embedded within statements. A Knowledge Graph (KG) is then constructed from these triplets where nodes are entities and arcs represent the relationships between nodes. In our algorithm, a relation extraction method is used to extract triplets from the statement we aim to verify the truthfulness of. We call this statement and its corresponding triplets as "a lay statement" and "lay triplets." The truthfulness of each lay triplet is verified based on an inference method corresponding to KG constructed from reliable sources.

After identifying the truthfulness of lay triplets, our algorithm additionally provides supporting evidence. Determining evidence for true triplets is relatively straightforward compared to identifying the evidence of falseness. Considering a true triplet, a supporting evidence is a set of paths between the subject and object entities inferred from KG associated with reliable sources. On the other hand, it is unclear how to obtain evidence for false triplets. Reasonable evidence for each false triplet should be a collection of relevant triplets extracted from KG under a specific condition. We explain our key idea with an example. Consider the false triplet ("property", "has_a", "space rocket"). We find in KG all triplets ("property", "has_a", $\bar{o}$). In this case, a set of all possible candidates $\bar{o}$ denoted as $\bar{O}$ can be {"bedroom," "kitchen," "bathroom," "roof," "garden," "shed," "swimming pool"}. A long proof of evidence can be this candidate evidence set and the fact that "space rocket" $\notin \bar{O}$. The drawback here is that the size of $\bar{O}$ can be very large. Summarizing the candidate collection into a concise but meaningful evidence set is challenging especially when the size of the collection is large.

In order to overcome this difficulty, we develop a novel algorithm to extract supporting evidence from concepts in ontologies. For any false triplet, we rely on the idea of representing each candidate with its broader concepts in ontologies given that the false triplet concept is not part of these broader concepts. Considering our example, an ontology could provide us with the fact that the first four terms in $\bar{O}$ are related to "house" and the remaining terms correspond to "backyard." Finally, as an evidence we provide ("property", "has_a", "house") and ("property",

"has_a", "backyard") and the facts "space rocket" is not "house," "space rocket" is not "backyard." Given the false triplet and its candidates, matching concepts in ontologies are considered. Then, we gather a set of potential evidence which includes candidate concepts and their broader concepts (satisfying some conditions). Evidence of various levels of granularity is constructed by a graph based algorithm on the subsumption tree of the ontology. We select an optimal collection of evidence from the potential evidence set. The optimized set of evidence is the smallest subcollection of the potential evidence set under the assumption that all candidates $\bar{o}$ have to be covered by themselves or their broader concepts which leads to a set covering problem.

In the rest of the paper, we consider the following running example. Given a false triplet ("Google", "OfficeLocationInUS", "Minneapolis"), we generate the evidence of falseness based on its relevant triplets from KG. The relevant triplets retrieved from KG have the "OfficeLocationInUS" relation associated with the "Google" or "Minneapolis" entity. In particular, we first find locations of Google's offices such as "Atlanta," "Chicago," "Los Angeles," "Miami," "Mountain View," etc. Also, companies whose office is located in Minneapolis such as "Target Corporation," "U.S. Bancorp," "Xcel Energy" are considered. These retrieved entities are used as the falseness evidence as we claim that "Minneapolis" is not part of all retrieved locations and similarly "Google" is not part of the set of retrieved companies. We rely on knowledge from ontologies to generate a concise set of evidence. For example, an ontology about geography is used to state that "Google" has offices in many states across U.S. while "Minneapolis" is located in Minnesota which is not one of these states.

Our main contribution is to provide supporting evidence for a given lay triplet after its truthfulness has been identified. If the triplet is true, then paths in KG provide evidence, however if false, then it is much more challenging to come up with the concept of evidence. To the best of our knowledge, no prior work provided supporting evidence of any given false lay statement by taking into account KG and ontologies. Our proposed system which combines knowledge from ontologies with predicate triplets from a KG contributes in this space. We specifically focus on selecting a complete set of falseness evidence to be as concise as possible. Also, our system relies mainly on both KG and ontologies which are constructed from reliable sources instead of knowledge from unverified web pages.

Our algorithm to provide supporting evidence along with the truthfulness of the lay triplet is applicable in various domains such as politics, sciences, news, and health care. Our work focuses on the health care domain as a case study mainly because of abundant health-related information available online and the importance of information quality. Specifically, a large number of medically related web sites are easily accessible online but only half of these sites have content reviewed by professionals (Gottleb, 2000). In addition, distorted information related to health conditions potentially causes devastating effects.

We summarize the literature in Section 2. In Section 3, we describe relevant background information, problem definitions, and thoroughly discuss our main algorithm. Data preparation and results of the algorithm based on our case study are reported in Section 4 while further discussions are provided in Section 5. Conclusion and future work are stated in Section 6.

## 2.RELATED WORK

Our algorithm verifies the truthfulness of any lay statement based on a KG thus we survey prior work in truth discovery-related fields. Substantial work exists in truth discovery for determining the veracity of multi-source data. In particular, the truth discovery problem aims to identify whether assertions claimed by multiple sources are true or false. Reliability of sources is also determined. Waguih and Berti (Waguih and Berti-Equille, 2014) provide an extensive review and an in-depth evaluation of 12 truth discovery algorithms. Additional truth discovery methods are proposed varying in many aspects to jointly estimate source reliability and truth statements (Li et al., 2014; Ma et al., 2015; Meng et al., 2015; Mukherjee et al., 2014; Xiao et al., 2016; Zhao et al., 2014; Zhi et al., 2015). These methods rely on a common assumption that information provided by a reliable source tends to be more trustworthy and the source providing trustworthy information is likely to be more reliable.

TruthOrRumor, a web-based truth judgment system, determines the truth based on results from a search engine (Liu et al., 2014). It considers reliability of data sources based on historical records and the copying relationship. Also, it implements currency determination techniques to take into account out-of-date statements. Wang et al. (2013) propose an algorithm to determine the truthfulness of a given statement based on a combination of a support score and credibility ranking value. While the support score measures how a web search result supports the statement, the credibility ranking computes the reliability of web pages. The t-verifier system (Li et al., 2011) requires users to pre-determine specific parts of statements to be verified. These systems take into account additional information of a data set or its source when determining the truthfulness of the statement.

Yin and Tan aim to distinguish true from false statements given a small set of ground truth facts (Yin and Tan, 2011). A graph optimization method is used in (Yin and Tan, 2011) where each node in the graph represents a statement and each edge connects a pair of relevant statements. Statements in the set of ground truth facts are labeled as 1. The algorithm assigns a truthfulness score ranging from -1 to 1 to each unlabeled statement. The scores of unlabeled statements not directly related to any labeled statements are possibly close to 0. This implies that the truthfulness of these statements remains undefined. Yamamoto and Tanaka propose a system to determine the credibility of a lay statement and extract aspects necessary to verify the factual validity from web pages (Yamamoto and Tanaka, 2009) whenever the statement is true. In order to estimate validity of a lay statement, the system collects comparative fact candidates using a web search engine. Fact candidates are sentences retrieved from the search engine that match a pattern specified by the lay statement. Then the validity of each candidate is computed based on the relation between the pattern and the entity contained in the candidate.

In comparison to the previous work, a focus of our algorithm is to provide concise but reliable supporting evidence in addition to identifying the truthfulness of a lay statement. The algorithm proposed by Yamamoto and Tanaka is similar to our system when the statement is true. In particular, both (Yamamoto and Tanaka, 2009) and our work use comparative candidate facts in order to assess the credibility of any lay statement. Instead of using web search engines, we rely on an inference method with respect to a KG to collect candidate triplets. A truthfulness score for the lay triplet is computed and compared against scores from those candidate triplets in order to determine whether the lay triplet is true or false. None of these works provide evidence of false statements which is the main contribution of our work.

## 3. METHODOLOGY

Content commonly found in textual documents especially online texts can be unreliable. In this study, we aim to identify whether a given lay statement is true or false and provide supporting evidence. We collect lay statements from many web pages publicly available online. We use a relation extraction algorithm (Rindflesch and Fiszman, 2003) to extract triplets consisting of subject entity $s$, object entity $o$ and their relation $r(s, o)$ embedded within lay statements. Our problem is scoped down to identifying the truthfulness of triplets representing the original lay statements. Knowledge obtained from reliable sources is an important factor in determining the trustworthiness of the lay triplets. We assure that the reliable resources are structured in a form of triplets $(s, r, o)$ which are used to construct a knowledge graph. Nodes and edges in KG represent entities and their relations, respectively. We write $(s, r, o) \in$ KG to mean that $s, o$ are nodes in KG and $r$ corresponds to an edge between them.

An evidence of falseness is obtained based on knowledge from various ontologies in related domains. In order to properly discuss falseness evidence and the main algorithm, we first provide a brief overview of a knowledge base (KB) or ontology and relevant background information. According to terminological knowledge, elementary descriptions are concept names (atomic concepts) and role names (atomic roles). Concept descriptions are built from concept and role names with concept and role constructors. All concept names and concept descriptions are generally considered as concepts. A deeper knowledge of ontologies can be obtained from (Franz et al., 2003).

Let KB = $(\mathcal{T}, \mathcal{A})$ be a knowledge base with $\mathcal{T}$ being a TBox and $\mathcal{A}$ an ABox as defined in (Franz et al., 2003). An interpretation $\mathcal{I} = (\Delta^{\mathcal{I}}, \cdot^{\mathcal{I}})$ is a model of KB corresponding to an ontology. We assume that KB is consistent. We assume that for each $(s, r, o) \in$ KG there are concepts $D, E$ in KB such that $s = D^{\mathcal{I}}$, $o = E^{\mathcal{I}}$. We denote $s = D^{\mathcal{I}}$ if and only if $D = C(s)$, where $D$ is a concept, i.e., given entity $s \in$ KG, $C(s)$ is the corresponding concept. We define special concepts $\top$ and $\bot$ as top (universal) and bottom (empty) concepts. Concept constructors such as an intersection ⊓, a union ⊔, and a negation ¬ combined with concept names are used to construct other concepts. Let $V_C$ be the set of all concept names. We also define $a \not\sqsubseteq b$ for concepts $a$ and $b$ if and only if $\exists y, y \neq \bot$ where y is a concept such that $y \sqsubseteq a \sqcap \neg b$.

A KB classification algorithm computes a partial order ≤ on a set of concept names with respect to the subsumption relationship, that is, $A \leq B \Leftrightarrow A \sqsubseteq B$ ($A$ sub-concept of $B$) for concept names $A$ and $B$. The classification algorithm incrementally constructs a graph representation in a form of a direct acyclic graph, called the subsumption tree, of the partial order induced by KB (Baader et al., 1994). Note that in this paper we use the term "tree" to use the term consistent with past literature. The underlying structure is actually an acyclic graph. Given $X$ as a set of concepts, computing the representation of this order is equivalent to identifying the precedence relation ≺ on $X$, i.e., $a \prec b$ for $a, b \in X$ if and only if $a \sqsubseteq b$ and if there exists $z \in X$ such that $a \sqsubseteq z \sqsubseteq b$, then $z = a$ or $z = b$.

Given the precedence relation $\prec_i$ for $X_i \subseteq X$, the incremental method defined in (Baader et al., 1994) computes $\prec_{i+1}$ on $X_{i+1} = X_i \cup \{c\}$ for some element $c \in X \setminus X_i$. The method consists of two main parts which are a top and a bottom search. The top and the bottom search identify sets $X_i \downarrow c = \{x \in X_i | c \sqsubseteq x \text{ and } c \not\sqsubseteq y \text{ for all } y \prec_i x, y \in X_i\}$ and $X_i \uparrow c = \{x \in X_i | x \sqsubseteq c \text{ and } y \not\sqsubseteq c \text{ for all } x \prec_i y, y \in X_i\}$. At the $i^{th}$ iteration, arcs corresponding to ≺ between $c$ and each element in $X_i \downarrow c$ as well as $c$ and each element in $X_i \uparrow c$ are added. Also, some existing arcs between elements in $X_i \downarrow c$ and $X_i \uparrow c$ are eliminated. At the end we have $a \prec b$ if and only if there is an arc in the constructed subsumption tree.

Our proposed system is built as a pipeline involving two main steps. We denote by $(s, r, o)$ the lay statement triplet that requires evidence.

1. Determining the truthfulness of the triplets: We rely mainly on the inference method called the Path Ranking Algorithm (PRA) introduced by Lao et. al. (2011) to verify whether each triplet $(\bar{s}, \bar{r}, \bar{o})$ in KG is true or false. The PRA produces $score_{PRA}$ for every pair of nodes. A PRA model is trained at each relation level. Particularly, the PRA model for a relation type $\bar{r}$ is trained to retrieve other nodes which potentially have a relation $\bar{r}(\bar{s}, \cdot)$ given node $\bar{s}$. We retrieve $\tilde{o}$ related to $(\bar{s}, \bar{r}, \tilde{o})$ with $score_{PRA}(\tilde{o}; \bar{s}) \geq \varepsilon_1$. All such object candidates $\tilde{o}$ are denoted by $\bar{O} = \bar{o}(\bar{s})$. A subject candidate set $\bar{S}$ is extracted in a similar way. The triplet $(s, r, o)$ is labeled as "True" if $score_{PRA}(s; o) \geq \varepsilon_2$ or $score_{PRA}(o; s) \geq \varepsilon_2$, and "False" otherwise. In addition, paths corresponding to high PRA scores are provided as supporting evidences of truthfulness if $(s, r, o)$ is true.

2. Extracting the evidence of falseness: We now assume that $(s, r, o)$ has been labeled as False in step 1, and either $s \notin \bar{S}$ for extracting a subject evidence of falseness or $o \notin \bar{O}$ for extracting an object evidence of falseness. Set $\bar{O}$ is the set of all objects that verify $s$ and $r$. If $(s, r, o)$ is false, then it has to be the case that $o \notin \bar{O}$ as otherwise $(s, r, o)$ would be true. Same holds for $\bar{S}$. We only discuss in detail the object evidence while the subject evidence is defined similarly.

Our validation for false statements do not rely on PRA, i.e. any inference algorithm on KG can be used. We found PRA to work best on our data. We next formally define evidence for false triplets. Recall that $\bar{O}$ is the set of all objects $\bar{o}$ for which $(s, r, \bar{o}) \in$ KG. It is important that these are all. In essence as a proof of falseness we can provide $\bar{O}$ together with the fact o $\notin \bar{O}$. However, this in many cases would provide a very long evidence since $|\bar{O}|$ is typically large. Instead we want to "aggregate" $\bar{O}$ into some smaller set $\alpha$ and still claim that o $\notin \alpha$. Wrapping these intuitions in the ontology formalism yields the following definition.

**Definition 1:** An object evidence of falseness is a collection $\alpha = \{\alpha_1, \alpha_2, \ldots, \alpha_k\}$ of concept names in KB such that

1) for each $\bar{o} \in \bar{O}$ there exists $i \in \{1, \ldots, k\}$ such that $C(\bar{o}) \sqsubseteq \alpha_i$,
2) there exists a concept $y, y \neq \bot$ such that $y \sqsubseteq C(o) \sqcap \neg(\bigsqcup_i \alpha_i)$ for all $i = 1, \ldots, k$.

The second condition can be rewritten as $C(o) \not\sqsubseteq \alpha_1 \sqcup \ldots \sqcup \alpha_k$ which in turn is equivalent to $C(o) \not\sqsubseteq \alpha_1$ and…and $C(o) \not\sqsubseteq \alpha_k$. In words, the second condition is equivalent to the requirement that $C(o)$ is not part of any element in the evidence set $\alpha$. The collection $\alpha$ is considered as an aggregated set of $\bar{O}$. We further define "potential evidence" $\alpha_i$ if there exist object evidence of falseness $\alpha$ such that $\alpha_i \in \alpha$.

From the definition, $\bar{O}$ is a set of object candidates having a relation $r(s, \bar{o})$ for the given triplet $(s, r, o)$. The first requirement in Definition 1 assures that each candidate $C(\bar{o})$ has to be subsumed by at least one potential evidence $\alpha_i$ ($\alpha$ is an aggregation of all elements in $\bar{O}$). As an example, letting $\alpha = \bar{O}$ satisfies the first requirement as each $C(\bar{o})$ for $\bar{o} \in \bar{O}$ is always subsumed by itself. According to the second requirement, concept $C(o)$ is not subsumed by any potential evidence $\alpha_i$. This ensures that concept $C(o)$ obtained from the false triplet $(s, r, o)$ does not belong to the evidence collection $\alpha$ (it mimics o $\notin \alpha$). Among all collections $\alpha$ we want to find the smallest one which is formalized later.

Referring to the false triplet example ("Google", "OfficeLocationInUS", "Minneapolis"), we let $\bar{O}$ be all object candidates retrieved from KG which have the relation

"OfficeLocationInUS" associated with subject "Google." Given $\bar{O}$ = {"Ann Arbor," "Atlanta," "Austin," "Birmingham," "Boulder," "Cambridge," "Chapel Hill," "Chicago," "Irvine," "Kirkland," "Los Angeles," "Miami," "Mountain View," "New York," "Pittsburgh," "Playa Vista," "Reston," "San Bruno," "San Francisco," "Seattle," "Sunnyvale," "Washington DC"}, selecting $\alpha = \bar{O}$ satisfies both requirements in Definition 1. The second requirement is satisfied as $C(o)$ associated with $o =$ "Minneapolis" is not subsumed by any element in the evidence collection $\alpha$. Moreover, the collection {"West region," "Northeast region," "South region," "Michigan," "Illinois"} is an example of a smaller evidence set which satisfies both requirements.

We propose Algorithm 1 to extract the evidence of falseness as defined in Definition 1. It is based on the subsumption tree (originally defined only for concept names $V_C$) which is expanded with negation concepts $V_{NC}$ and specific concepts $V_F$. The set $V_{NC}$ is formally defined as $V_{NC}$ = { $\neg v \mid v \in V_C$}. Recall that we define $a \not\sqsubseteq b$ for concepts $a$ and $b$ if and only if $\exists y, y \neq \bot$ where y is a concept such that $y \sqsubseteq a \sqcap \neg b$ which involves $\neg b$ and mandates $V_{NC}$. Even though infinitely many concepts can be constructed from concept names and concept constructors, we only focus on specific concepts $V_F$ which ensure the second requirement in Definition 1. A concept $f \in V_F$ corresponds to $f = x \sqcap \neg c$ for $c \in V_C, x \in V_C$ and no proper concept name or concept name negation between $\bot$ and $f$. Algorithm 1 extracts potential evidence by considering nodes along paths in the tree which satisfy both requirements in Definition 1. In order for Algorithm 1 to check the satisfiability of the requirements, not only $V_C$ but also both $V_{NC}$ and $V_F$ have to be included in the subsumption tree. Hence, the standard tree consisting of concept names $V_C$ only has to be expanded. An algorithm to add $V_{NC}$ and $V_F$ to the existing standard tree is provided in Appendix A. The subsumption tree used in Algorithm 1 is of the form $\mathcal{G} = (V_C \cup V_{NC} \cup V_F, A)$ where $\mathcal{G}$ is a directed acyclic graph with root ⊤ (top concept). We also define paths and nodes associated with the subsumption tree used in Algorithm 1 as follows.

**Definition 2:** $Path_{ij}$ is a set of all possible paths from node $j$ to node $i$ in the subsumption tree. For $path\ P \in Path_{i\top}$ we denote by $P_m$ the $m^{th}$ node in $P$ starting from ⊤. Let $Node_{ij}$ be the set of all nodes along all paths in $Path_{ij}$.

**Algorithm 1** ($o$, $\bar{O}$) with $\mathcal{G} = (V_C \cup V_{NC} \cup V_F,\ A)$:

---

1  Set $sup_{C(o)} = \emptyset$
2  For each $\bar{o} \in \bar{O}$:
3      Set $sup_{C(\bar{o})} = \emptyset$
4      For each $path\ P \in Path_{C(\bar{o})\top}$:
5          For $m = length(P)$ to 1:
6              If $P_m \in V_C$:
7                  $\Omega_{m,P} = \{y \in V_C \cup V_{NC}|\ y \in Node_{\bot C(o)} \cap Node_{\bot \neg P_m},\ y \neq \bot\}$
8                  If $\Omega_{m,P} \neq \emptyset$:
9                      Add $P_m$ to $sup_{C(\bar{o})}$
10                 Else:
11                     Break
12     $sup_{C(o)} = sup_{C(o)} \cup sup_{C(\bar{o})}$
13 Remove duplicate nodes in $sup_{C(o)}$
14 Return $\alpha$ = SetCover ($sup_{C(o)}$)

---

In Algorithm1, we assume that $\exists y \in V_C \cup V_{NC}, y \neq \bot$ such that $y \sqsubseteq C(o) \sqcap \neg C(\bar{o})$ for all $\bar{o} \in \bar{O}$ and $o$ in KB. This assumption implies that $C(o)$ cannot be part of $C(\bar{o})$ for an element in $\bar{O}$. If $C(o) \sqsubseteq C(\bar{o})$ for an $\bar{o} \in \bar{O}$, then the statement is true. It assures that $sup_{C(\bar{o})}$ incremented in step 9 for each $\bar{o} \in \bar{O}$ is not empty.

Algorithm 1 repeats steps 3-11 to compute potential evidence $\alpha_i$'s for each $\bar{o} \in \bar{O}$ and stores them in $sup_{C(o)}$. All $sup_{C(\bar{o})}$'s are combined in $sup_{C(o)}$ according to step 12. The set $sup_{C(o)}$ is equivalent to the set of all potential evidences. Note that for every $a \in sup_{C(o)}$ there is $\bar{o}$ such that $P \in Path_{C(\bar{o})\top}$ contains $a$ due to steps 2-12. Algorithm 1 is specifically constructed to ensure that every element in $sup_{C(o)}$ is one of nodes along at least one path from $C(\bar{o})$ to the root. This implies that elements in $sup_{C(o)}$ can be considered as broader concepts of candidate evidence $\bar{o} \in \bar{O}$.

The first requirement in Definition 1 requires that any $\bar{o} \in \bar{O}$ has at least one corresponding $\alpha_i$ that subsumes $C(\bar{o})$. Hence, the algorithm considers nodes along all possible paths from $C(\bar{o})$ to the root (top concept ⊤) for every $\bar{o} \in \bar{O}$ in order to extract potential evidence $\alpha_i$. The second requirement in Definition 1 is directly associated with Ω computed and verified in steps 7 and 8. Both $V_{NC}$ and $V_F$ in the subsumption tree used in Algorithm 1 are necessary to compute Ω, i.e., $V_{NC}$ and $V_F$ guarantee that $y \in V_C \cup V_{NC} \cup V_F$. Particularly, $\Omega_{m,P} \neq \emptyset$ implies that $C(o) \not\sqsubseteq P_m$; therefore, $P_m$ in this case can be considered as potential evidence $\alpha_i$. Algorithm 1 then computes $\Omega_{m,P}$ for each node $P_m \in V_C$ in path $P \in Path_{C(\bar{o})\top}$ corresponding to each $\bar{o} \in \bar{O}$. If $\Omega_{m,P}$ is not empty, $P_m$ (considered as potential evidence $\alpha_i$) is added to $sup_{C(\bar{o})}$ in step 9. Elements in $sup_{C(\bar{o})}$ correspond to nodes in $V_C$ and thus to concept names. They also correspond to potential evidence $\alpha_i$'s. Each $\Omega_{m,P}$ computed in Algorithm 1 considers concept names and negation of concept names but Definition 1 considers any concept. Algorithm 1 consequently provides an approximate evidence set while an exact algorithm is discussed later.

Figure 1 illustrates Algorithm 1. The path from $C(o)$ to the root ⊤ is highlighted in blue. Nodes along red paths are collected as potential evidence $\alpha_i's$ as $C(o)$ is not subsumed by these nodes ($\Omega$ corresponding to these nodes are not empty).

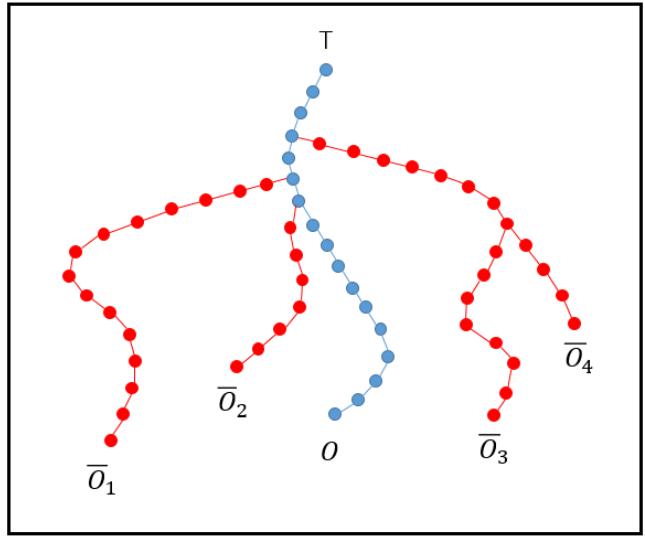

Figure 1: Illustration of Algorithm 1

Regarding the run time analysis, the proposed algorithm consists of nested loops in steps 2, 4 and 5. Let $N$ be the number of nodes in $\mathcal{G}$ and $M$ the maximum number of paths between any node and the root ⊤. The most outer loop in step 2 considers each element $\bar{o} \in \bar{O}$ which is O($N$) while the middle loop in step 4 processes each path $P \in Path_{C(\bar{o})\top}$ corresponding to $\bar{o}$ in step 2, i.e., O($M$). Also, each node along all paths from $C(\bar{o})$ to the root is considered in the most inner loop in step 5, which is O($N$). Computing $\Omega_{m,P}$ for each node $P_m$ requires O($N^2$). Hence, the computational complexity of Algorithm 1 is O($N^4 \cdot M$). Algorithm 1 can be sped up by using bisection. The more efficient version is provided in Appendix B.

Referring to the running example, we let $o =$ "Minneapolis" and consider $\bar{o} =$ "Mountain View." We consider paths from $C$("Mountain View") to the root as well as all nodes along these paths. Node $P_m = C(\text{"Mountain View"})$ is added to $sup_{C(\bar{o})}$ as $Node_{\perp C(\text{"Minneapolis"})} \cap Node_{\perp \neg C(\text{"Mountain View"})}$ is not empty. Particularly, there exists a node which belongs to both sets, i.e., $C(\text{"Minneapolis"}) \in Node_{\perp C(\text{"Minneapolis"})}$ and $C(\text{"Minneapolis"}) \in Node_{\perp \neg C(\text{"Mountain View"})}$. According to a natural geography ontology associated with the example, $sup_{C(\bar{o})} = \{C(\text{"Mountain View"}), C(\text{"Santa Clara"}), C(\text{"California"}), C(\text{"West region"})\}$ is retrieved. Note that $C(\text{"USA"})$ is not in $sup_{C(\bar{o})}$ since $Node_{\perp C(\text{"Minneapolis"})} \cap Node_{\perp \neg C(\text{"USA"})}$ is empty. Intuitively, "Minneapolis" is a location in "USA" ("Minneapolis" is part of "USA") and therefore, "USA" cannot be counted as an evidence of falseness.

After obtaining the set of all potential $\alpha$ across all possible evidences, we aim to compute an optimal set of evidence with the smallest cardinality. We formally define the object evidence of falseness problem $EP$ as $Z_{EP} = \min_{\alpha \text{ object evidence of falseness}} |\alpha|$. A set covering problem is proposed to find an optimal set of evidence. We later give a condition when it solves it optimally. The set covering problem is formulated as follows.

SetCover($sup_{C(o)}$):

Universe $U = \{\bar{o}_1, \bar{o}_2, \ldots, \bar{o}_{|\bar{O}|}\}$

For any node $a \in sup_{C(o)}$, we define $T_a = \{\bar{o} \in U | a \in P \text{ where } P \in Path_{C(\bar{o})\top}\}$

The set covering problem $SC$ reads $Z_{SC} = \min |I|$ subject to $\bigcup_{i \in I} T_i = U$. For any node $a \in sup_{C(o)}$, we know that $T_a \subseteq U$ which is necessary for the feasibility of $SC$. The set covering problem aims to find a minimum number of set $T_a$'s for $a \in sup_{C(o)}$ so that selected sets contain all elements in the universe $U$, i.e. they cover $\bar{O}$. A feasible solution to the set covering problem satisfies the first requirement of Definition 1. The set $T_a$ for each $a \in sup_{C(o)}$ is specifically constructed based on $sup_{C(o)}$. Note that the set $T_a \neq \emptyset$ because of the fact that for every $a \in sup_{C(o)}$ there is $\bar{o}$ such that $P \in Path_{C(\bar{o})\top}$ contains $a$ and the construction of $T_a$. All elements in $sup_{C(o)}$ added in Algorithm 1 are guaranteed to satisfy the second requirement of Definition 1.

According to the false triplet ("Google", "OfficeLocationInUS", "Minneapolis") example, we consider $o =$ "Minneapolis" and the set $\bar{O}$ given previously. A set of generated $T_a's$ which yields a feasible solution to the set covering problem is given in Table 1. The left column lists 5 elements from $sup_{C(o)}$.

Table 1. An example of feasible $T_a's$ sets for the set covering problem

| | |
|---|---|
| $T_{C(\text{"West region"})}$ | "Boulder," "Irvine," "Kirkland," "Los Angeles," "Mountain View," "Playa Vista," "San Bruno," "San Francisco," "Seattle," "Sunnyvale" |
| $T_{C(\text{"Northeast region"})}$ | "Cambridge," "New York," "Pittsburgh" |
| $T_{C(\text{"South region"})}$ | "Atlanta," "Austin," "Chapel Hill," "Miami," "Reston," "Washington DC" |
| $T_{C(\text{"Michigan"})}$ | "Ann Arbor," "Birmingham" |
| $T_{C(\text{"Illinois"})}$ | "Chicago" |

Propositions 1 and 2 stated next establish the relationship between $EP$ and $SC$. Proofs of Propositions 1 and 2 are provided in Appendix C.

**Proposition 1:** $SC$ is feasible and a feasible solution to $SC$ yields a feasible solution to $EP$ of same or smaller cardinality.

This implies that $Z_{SC} \geq Z_{EP}$. Due to Proposition 1, a solution to $SC$ is always a feasible solution to $EP$ and thus an object evidence of falseness obtained from $SC$ can be used as a representative of the evidence set from $EP$.

We consider either concept names or negation of concept names when $\Omega_{m,P}$ is computed in Algorithm 1. An exact algorithm replaces $\Omega_{m,P}$ defined in step 7 of Algorithm 1 with $\Omega'_{m,P} = \{$concept $y \mid y \sqsubseteq C(o) \sqcap \neg P_i, y \neq \bot\}$. All concepts constructed from concept names and concept constructors are considered in $\Omega'_{m,P}$. We also observe that checking $\Omega'_{m,P} \neq \emptyset$ is equivalent to checking satisfiability of the concept $C(o) \sqcap \neg P_m$, i.e. if $C(o) \sqcap \neg P_m$ is satisfiable, then $\Omega'_{m,P} \neq \emptyset$ as stated in (Baader, 2003).

**Proposition 2:** If $\Omega_{m,P}$ in step 7 of Algorithm 1 is substituted with $\Omega'_{m,P}$, then $Z_{SC} = Z_{EP}$.

In the proof for Proposition 2, we show that a feasible solution to $EP$ is also a feasible solution to $SC$ when replacing $\Omega_{m,P}$ with $\Omega'_{m,P}$. This implies that $Z_{EP} \geq Z_{SC}$ and combined with Proposition 1 it yields $Z_{SC} = Z_{EP}$.

We define a subject evidence $\beta = \{\beta_1, \beta_2, \ldots, \beta_k\}$ in the same way. In order to identify the subject evidence of falseness $\beta$, Algorithm 1 is applied where all definitions and propositions are defined similarly with respect to $(\bar{s}, r, o)$. A domain under consideration can have multiple ontologies. In such a case, we implement the proposed algorithms to identify the evidence of falseness for each ontology. The minimum cardinality of subject/object evidence is selected across all ontologies. Finally, the problem to identify the evidence of falseness for each triplet $(s, r, o)$ is formulated by considering both subject and object evidence sets as

$$\min\left\{\min_{ont\,\in\,\text{ontologies}}\left(\min_{\alpha\text{ object evidence in }ont}|\alpha|\right)_{ont},\min_{ont\,\in\,\text{ontologies}}\left(\min_{\beta\text{ subject evidence in }ont}|\beta|\right)_{ont}\right\}.$$

## 4. CASE STUDY

We apply the proposed algorithm to the health care domain as a case study. A reliable source in our case is obtained from biomedical publications stored in the MEDLINE database. In order to construct KG, SemRep (Rindflesch and Fiszman, 2003) is used to extract semantic predicate triplets from biomedical texts. SemRep matches subject and object entities in triplets with concepts from the UMLS Metathesaurus and matches relationship with respect to the UMLS Semantic Network. It also takes into account a syntactic analysis, a structured domain knowledge, and hypernymic propositions extensively. The data contains both the extracted triplet and the corresponding sentence from MEDLINE.

We first train the PRA model based on KG constructed from SemRep. We further compare its performance with the evaluation metrics reported in (Lao et al., 2011) where PRA has been trained on the NELL data set. The average mean reciprocal rank (MRR) across different relation types reported in (Lao et al., 2011) is 0.516 while the average MRR of PRA on SemRep is 0.25. The MRR is computed based on the rank of the first correctly retrieved triplet; however we aim to correctly retrieve all triplets that are in KG. As a result, we additionally compute the mean average precision (MAP) which considers the rank position of each triplet in KG. The MAP based on our trained PRA model is 0.1. This implies that on average every 10th retrieved result is correct. We then manually inspect the original statements and their corresponding predicate triplets extracted from SemRep. Even though a preliminary evaluation of SemRep reported in (Rindflesch and Fiszman, 2003) states 83% precision, extracted predicate triplets in KG contain many errors based on our manual observation. Examples of predicate triplets incorrectly extracted from original statements are provided in Appendix D. The issue is that the sentences are clearly correct but the extracted triplets are often wrong.

Hence, we pre-process KG by verifying each extracted predicate triplet with the PRA model and additional relation extraction systems. Detailed explanations are provided in the following data preparation section.

**4.1 Data Preparation**

We aim to re-construct KG containing only triplets with high precision. After manually observing results from the trained PRA model, triplets with high PRA scores tend to be more accurate than those with low PRA scores. Hence, PRA is one of models used to verify triplets in KG. We further employ other relation extraction systems to filter out incorrect triplets from the original KG. Ollie (Mausam et al., 2012) is an open information extraction software which aims to extract binary relationships from sentences. According to open information extraction, a schema of relations does not need to be pre-specified. In addition, we train a recurrent neural network model called LSTM-ER proposed by Miwa and Bansal (Miwa and Bansal, 2016) on a publicly available training data set having gold standard labels. Each instance in the training data consists of a statement and its predicate triplet. The training data set used to train the LSTM-ER model includes the ADE corpus (Gurulingappa et al., 2012), SemEval-2010 (Hendrickx et al., 2010), BioNLP (Kim et al., 2011), and the SemRep Gold standard annotation (Kilicoglu et al., 2011).

In order to pre-process the original KG, we propose a strategy to combine triplets with high PRA scores and triplets matching with the Ollie or LSTM-ER models. A flow diagram of the proposed strategy in order to construct an adjusted KG is depicted in Figure 2.

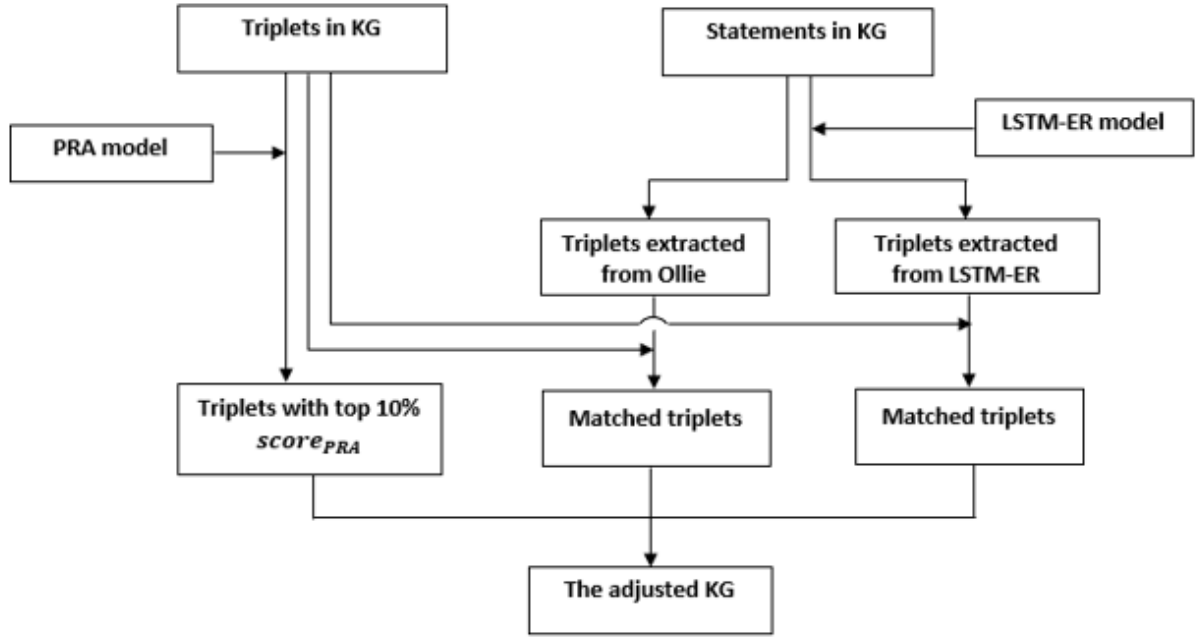

Figure 2: The flow diagram of the strategy to pre-process KG

According to the proposed strategy, we infer the trained PRA model on KG and rank the results from high to low PRA scores. The triplets positioned in the top 10 percent of the ranked PRA scores are retrieved. Additionally, we collect all possible matches between triplets in KG and results from Ollie. We also use the trained LSTM-ER model to infer possible relations from statements associated with triplets in the original KG. Each triplet from KG is collected if its relation matches with the relation inferred from the LSTM-ER model. We conduct a preliminary experiment by extracting matched triplets using Ollie, and the LSTM-ER model based on 10,000 randomly selected triplets. Based on the experiment, we observe a small proportion of matching triplets among different relation extraction models. A detailed discussion of the experiment is provided in Appendix E.

Additionally, we observe that many statements in the original KG involve studies with non-human subjects such as "Effects of acetylcholine, histamine, and serotonin infusion on venous return in dogs." In order to filter out these statements, we consider the UMLS semantic type, a categorization of concepts represented in the UMLS Metathesaurus, tagged in the statements. In particular, we eliminate statements which contain "Amphibian," "Animal," "Bird," "Fish," "Mammal," "Reptile," and "Vertebrate" semantic types. We provide the number of nodes and edges in the original KG and in the adjusted KG in Table 2. The average MRR and the average MAP based on the PRA model on the adjusted KG are 0.44 and 0.29, respectively

Table 2: Number of nodes and edges in the original KG and the adjusted KG

|  | **The original KG** | **The adjusted KG** |
|---|---:|---:|
| **Number of nodes** | 229,063 | 161,930 |
| **Number of edges** | 15,700,435 | 4,107,296 |

**4.2 Results**

We run the whole pipeline of Algorithm 1 to validate the truthfulness and provide supporting evidence of lay triplets with the adjusted KG. Based on 2,084 lay triplets consisting of 20 relation types collected from health-related web pages, we identify the truthfulness of each triplet and extract evidence candidates as summarized in Appendix F. Across all relation types, there are 501 false triplets which account for 24 percent of all lay triplets.

Instead of directly specifying thresholds $\varepsilon_1$ and $\varepsilon_2$ in step 1 of the process, we identify the rank threshold $rank_{\varepsilon_1}$ and $rank_{\varepsilon_2}$ based on the ordered PRA scores. To identify $rank_{\varepsilon_1}$ corresponding to subject entity $s$ and relation type $r$, we retrieve a set of all object entities which have relation $r(s,\cdot)$ identified by the PRA model. This set is denoted as $O_{all}$. The set $O_{KG} = \{\tilde{o} \mid (s, r, \tilde{o}) \in KG\}$ is also retrieved. A parameter $x$ which is defined as $x = \mathbf{1}_{|O_{KG}|>0} \frac{|O_{KG}|}{median(PRA\ scores\ of\ O_{KG})}$ is used to specify the rank threshold $rank_{\varepsilon_1}$ as follows:

$$rank_{\varepsilon_1} = \begin{cases} 5 & \text{if } x \leq 0.25 \\ 10 & \text{if } 0.25 < x \leq 0.5 \\ 15 & \text{if } 0.5 < x \leq 0.75 \\ 20 & \text{if } x > 0.75 \end{cases} + 5 - \frac{|O_{all}|}{10000}.$$

The parameter $x$ captures how well the PRA model gives high ranks to triplets in KG. The higher the value of $x$ is, the better the PRA model performs. This implies that $rank_{\varepsilon_1}$ should vary proportionally to $x$. The middle term is a hyper parameter calibrated in the experiment in order to obtain the best performance. The last term in the $rank_{\varepsilon_1}$ formula takes into account how $|O_{all}|$ affects $rank_{\varepsilon_1}$. Having high $|O_{all}|$ indicates that many object entities are predicted to have $r(s,\cdot)$ with respect to subject $s$. Therefore, it is more challenging for the PRA model to correctly rank retrieved object entities. This implies that high $|O_{all}|$ leads to low value of the threshold $rank_{\varepsilon_1}$ as expressed in the formula. We extract entity $\bar{o}$ whose rank based on PRA score is higher than $rank_{\varepsilon_1}$ as candidates in $\bar{O}$. Moreover, we specify $rank_{\varepsilon_2}$ as $\max(rank_{\varepsilon_1}, 0.005 * O_{all})$ to identify the truthfulness of $(s, r, o)$. If the rank of $score_{PRA}(o; s)$ is higher than $rank_{\varepsilon_2}$, we specify $(s, r, o)$ as true.

Among 501 false lay triplets, we first eliminate triplets whose object $o$ does not match with concepts in ontologies. Candidates are then used to compute the evidence set based on the remaining 395 triplets by using Algorithm 1. We only perform evaluations on object candidates while subject candidates can be done similarly. The average cardinality of object candidates $|\bar{O}|$ and the average cardinality of their corresponding evidence sets $|\alpha|$ across all relation types are 11.65 and 2.24, respectively. A histogram of the produced object evidence $|\alpha|$ of all relation types based on candidates $\bar{O}$ is provided in Figure 3.

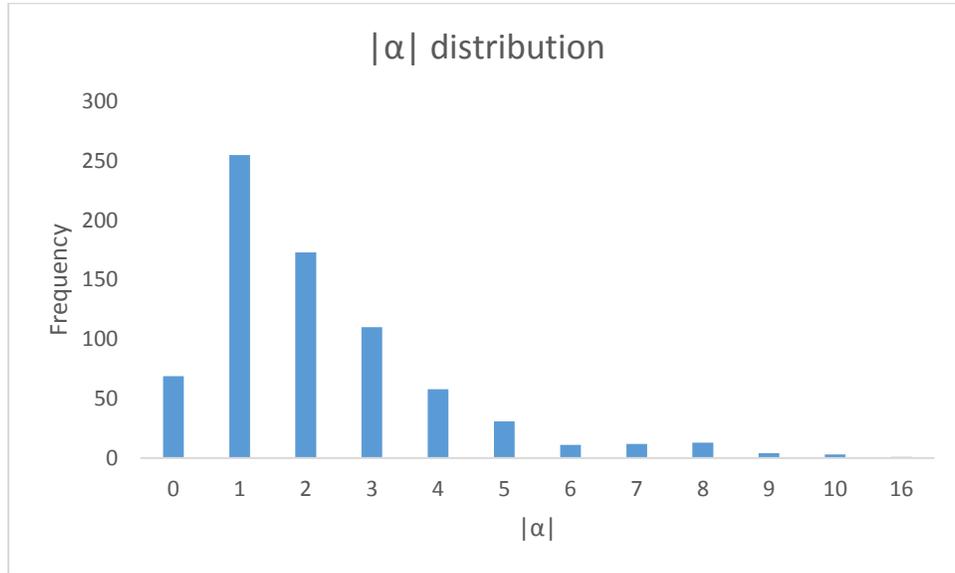

Figure 3: A histogram of object evidence $|\alpha|$

In order to evaluate the performance of the proposed algorithm, we choose 3 relation types representing high, medium and low MAP computed from the PRA model which are "TREATS," "DIAGNOSES," and "CAUSES," respectively. For each relation type, we select 5 cases to compare the evidence sets resulting from the algorithm (denoted as "Al") against the evidence sets constructed manually (denoted as "Ma") as illustrated in Table 3. A complete comparison of elements in evidence sets "Al" and "Ma" is provided in Appendix G.

Table 3: Evidence sets obtained from the algorithm and the evidence sets constructed manually

| False triplets | |Al| | |Ma| | |Al ∩ Ma| | |Al\Ma| | |Ma\Al| |
|---|---|---|---|---|---|
| **Relation type: TREATS** | | | | | |
| Heparin TREATS Fever | 1 | 1 | 1 | 0 | 0 |
| Amiodarone TREATS Hepatitis C | 1 | 1 | 1 | 0 | 0 |
| Stress management TREATS Mitral Valve Prolapse | 1 | 1 | 1 | 0 | 0 |
| Capoten TREATS Coughing | 1 | 1 | 1 | 0 | 0 |
| Losartan TREATS Varicose Ulcer | 3 | 2 | 1 | 2 | 1 |
| **Average of TREATS** | 1.4 | 1.2 | 1.0 | 0.4 | 0.2 |
| **Relation type: DIAGNOSES** | | | | | |
| Echocardiography DIAGNOSES Hyperlipidemia | 1 | 1 | 1 | 0 | 0 |
| Platelet Size DIAGNOSES Anemia | 1 | 2 | 1 | 0 | 1 |
| Esophageal pH Monitoring DIAGNOSES Malignant breast neoplasm | 5 | 3 | 3 | 2 | 0 |
| Cholesterol measurement test DIANOSES Malignant breast neoplasm | 5 | 2 | 2 | 3 | 0 |
| Electrocardiogram DIAGNOSES Muscle strain | 1 | 1 | 1 | 0 | 0 |
| **Average of DIAGNOSES** | 2.6 | 1.8 | 1.6 | 1.0 | 0.2 |
| **Relation type: CAUSES** | | | | | |
| Caffeine CAUSES Gout | 1 | 3 | 1 | 0 | 2 |
| hypercholesterolemia CAUSES Neuropathy | 2 | 2 | 2 | 0 | 0 |
| Leukemia CAUSES Gout | 2 | 2 | 2 | 0 | 0 |
| Harpin CAUSES Cardiomegaly | 2 | 1 | 1 | 1 | 0 |

| Ascorbic Acid CAUSES Senile Plaques | 2 | 2 | 1 | 1 | 1 |
|---|---|---|---|---|---|
| **Average of CAUSES** | 1.8 | 2.0 | 1.4 | 0.4 | 0.6 |
| **Average across 3 relation types** | 1.9 | 1.7 | 1.3 | 0.6 | 0.3 |

## 5.CONCLUSIONS AND FUTURE WORK

In this work, we develop a system to validate the truthfulness of lay triplets and provide supporting evidence. Our system employs the PRA algorithm inferred on KG re-constructed from reliable sources to identify whether a lay triplet is true or false. In our experiment, we train the PRA model based on KG constructed from biomedical literature. The original KG contains incorrect triplets due to the relation extraction process. We attempt to re-construct KG consisting of more accurate triplets by verifying each triplet in the original KG with additional relation extraction algorithms. The trained PRA model on the adjusted KG yields 0.44 MRR and 0.29 MAP averaged across all relation types. The performance of the PRA model based on the adjusted KG is improved. However, the adjusted KG still contains errors due to the challenge of complicated biomedical text and limited resources in training additional relation extraction algorithms.

We use a combination of knowledge from ontologies and triplets in the adjusted KG to extract a concise supporting evidence set. Specifically, Algorithm 1 aims to find the supporting evidence set which does not overlap with an entity in a lay triplet. The evidence set is aggregated from candidates obtained from triplets in the adjust KG by using knowledge of ontologies. We apply Algorithm 1 to extract evidence sets based on each ontology and repeatedly consider all possible ontologies. It is reasonable to select the ontology which yields the minimum cardinality of evidence sets. According to our algorithm, we first match object (subject) entity in a lay triplet with concepts within all ontologies. Non-matching ontologies are not taken into a consideration when object (subject) candidates are paired with concepts in ontologies. We assume that candidates that cannot be matched with concepts in the same ontology as the lay triplet's are disregarded.

We consider the number of candidates $|\bar{O}|$ extracted from the PRA model and compare it against the cardinality of the evidence set $|\alpha|$ resulted from the algorithm. The average of $|\bar{O}|$ is larger than the average of $|\alpha|$ by a factor of 5 across all relation types. This implies that our proposed algorithm provides valid and concise evidence sets. To evaluate the performance of our algorithm, we compare the evidence set extracted from our proposed algorithm with a manually-constructed evidence set. The average number of overlap between the evidence set from the algorithm and the manually constructed set is 74% across the 3 relation types. Our proposed algorithm performs very well especially with some specific relation types such as "TREATS" with the overlap of 87%.

The problem is challenging due to limited resources to construct a complete and accurate KG. An imperfect KG plays a significant role in the inferior performance of the PRA model which directly impacts the performance of Algorithm 1 to extract evidence sets. A better quality of KG would lead to a higher performance of the proposed system. Hence, future work should focus on improving relation extraction algorithms to construct KG. We believe that this is of utmost importance, not just for our work, but all systems that rely on SemRep.

# Appendix A

Algorithm A1 presented next adds negation nodes $V_{NC}$ to the existing subsumption tree by implementing the incremental methods defined in (Baader et al., 1994). An augmented tree is of the form $\mathcal{G} = (V_C \cup V_{NC} \cup V_F, A)$ where nodes $v \in V_C$ are concept names, nodes $v \in V_{NC}$ are negation concepts of concept names and nodes $v \in V_F$ are specific concepts which ensure the second requirement in Definition 1. Arc $(b, a) \in A$ between node $a$ and $b$ has the following properties.

1. $a \in V_C \cup V_{NC}$ and $b \in V_C \cup V_{NC}$ => $a \prec b$, i.e. $a \leq b$ and if there exists $z \in V_C \cup V_{NC}$ such that $a \leq z \leq b$, then $z = a$ or $z = b$

2. $a \in V_F$ and $b \in V_C \cup V_{NC}$ => $a \leq b$

Concept names and concept constructors are combined in order to construct concepts. This consequently yields a significant number of concepts. Nodes in $V_F$ may have $\leq$ relation with many concepts which are not added to the subsumption tree. Note that we only add necessary concepts required for Algorithm 1. Therefore, the first property only takes $V_C \cup V_{NC}$ into a consideration, i.e., we are not able to assume $\prec$ relation between nodes in $V_C \cup V_{NC}$ and nodes in $V_F$.

**Algorithm A1** to generate $\mathcal{G} = (V_C \cup V_{NC} \cup V_F, A)$:

---

1      Initially set $X = V_C$

2      For each concept name $v \in V_C$:

3          create node $c = \neg v$

4          compute set $X \downarrow c = \{x \in X \mid c \leq x \text{ and } c \nleq y \text{ for all } y \prec x, y \in X\}$

5          compute set $X \uparrow c = \{x \in X \mid x \leq c \text{ and } y \nleq c \text{ for all } x \prec y, y \in X\}$

| | |
|---|---|
| 6 | add arcs between $c$ and each element of $X \downarrow c$, and between $c$ and each element of $X \uparrow c$ |
| 7 | remove all arcs between elements of $X \uparrow c$ and $X \downarrow c$ |
| 8 | compute set $X^{V_C} \parallel c = \{x \in X, x \in V_C \mid \bot \leq c \sqcap x\}$ |
| 9 | For each element $d \in X^{V_C} \parallel c$: |
| 10 | create artificial node $a_d = c \sqcap d$ and add an arc $(a_d, \bot)$ connecting $\bot$ to $a_d$ |
| 11 | add an arc $(d, a_d)$ connecting $a_d$ to $d$ and $(c, a_d)$ connecting $a_d$ to $c$ |
| 12 | $X = X \cup \{c\}$ |

To extract the evidence set, Algorithm 1 computes $\Omega's$ in order to ensure the second requirement in Definition 1. Any set $\Omega_{m,P}$ depends on an overlap between two set of nodes corresponding to $C(o)$ and $\neg P_m$. To verify the overlap of these two sets, both $V_{NC}$ and $V_F$ are necessary. Algorithm A1 adds $V_{NC}$ and $V_F$ to the standard subsumption tree which only includes concept names $V_C$. It relies mainly on an incremental method involving the top and bottom search computed in steps 4 and 5. As arcs in the tree only represent the subsumption relationship, steps 8-11 further take into account an overlap case which involves nodes in $V_F$.

**Appendix B**

As a more efficient version of Algorithm 1, we propose Algorithm B1 using a bisection method based on $\Omega_{m,P}$ as of follows.

**Algorithm B1** $(o, \bar{O})$ with $\mathcal{G} = (V_C \cup V_{NC} \cup V_F, A)$:

| | |
|---|---|
| 1 | Set $sup_{C(o)} = \emptyset$ |
| 2 | For each $\bar{o} \in \bar{O}$: |
| 3 | Set $sup_{C(\bar{o})} = \emptyset$ |
| 4 | For each $path\ P \in Path_{C(\bar{o})^\top}$: |
| 5 | $m = \left\lfloor \frac{length(P)}{2} \right\rfloor$ |
| 6 | While True: |
| 7 | $\Omega_{m,P} = \{y \in V_C \cup V_{NC}|\ y \in Node_{\perp C(o)} \cap Node_{\perp \neg P_m}, y \neq \perp\}$ |
| 8 | If $\Omega_{m,P} \neq \emptyset$: |
| 9 | $\Omega_{m-1,P} = \{y \in V_C \cup V_{NC}|\ y \in Node_{\perp C(o)} \cap Node_{\perp \neg P_{m-1}}, y \neq \perp\}$ |
| 10 | If $\Omega_{m-1,P} = \emptyset$: |
| 11 | For $m' = m$ to $length(P)$: |
| 12 | Add $P_{m'}$ to $sup_{C(\bar{o})}$ |
| 13 | Break |
| 14 | Else: |
| 15 | $m = \left\lfloor \frac{m}{2} \right\rfloor$ |
| 16 | Else: |
| 17 | $m = m + \left\lfloor \frac{m}{2} \right\rfloor$ |
| 18 | $sup_{C(o)} = sup_{C(o)} \cup sup_{C(\bar{o})}$ |
| 19 | Remove duplicate nodes in $sup_{C(o)}$ |
| 20 | Return $\alpha = \text{SetCover}(sup_{C(o)})$ |

Note that as in Algorithm 1 $\forall a \in sup_{C(o)}$ there is $\bar{o}$ such that $P \in Path_{C(\bar{o})^\top}$ contains $a$ due to steps 2-18. For any $\bar{o}_i \in \bar{O}$, Algorithm B1 identifies the $m^{th}$ position in each path $P \in Path_{C(\bar{o}_i)^\top}$ such that $\Omega_{m,P} \neq \emptyset$ and $\Omega_{m-1,P} = \emptyset$ by the bisection search method. We know that $P_m \sqsubseteq P_{m-1} \Leftrightarrow \neg P_{m-1} \sqsubseteq \neg P_m$ for any concepts $P_m$ and $P_{m-1}$. Hence, $\Omega_{m-1,P} \neq \emptyset$ implies that $\Omega_{m,P} \neq \emptyset$. Step 12 adds all nodes which are subsumed by $P_m$ to $sup_{C(\bar{o}_i)}$ if conditions in steps 8 and 10 are satisfied

As in the run time analysis of Algorithm 1, we let $N$ be the number of nodes in $\mathcal{G}$ and $M$ the maximum number of paths between any node and the root $\top$. Similarly to Algorithm 1, the most outer loop in step 2 is O($N$) while the middle loop in step 4 is considered as O($M$). Algorithm B1 relies on the bisection search which is accounted for O($\log_2 N$). Computing $\Omega_{m,P}$ for each node $P_m$ requires O($N^2$). Therefore, the computational complexity of Algorithm B1 is O($\log_2 N \cdot N^3 \cdot M$) compared to O($N^4 \cdot M$) corresponding to Algorithm 1.

**Appendix C**

**Proof of Proposition 1:**

We first argue that $SC$ is feasible. For any path $P \in Path_{C(\bar{o}_i)\top}$ of $\bar{o}_i \in \bar{O}$, $P_{length(P)} = C(\bar{o}_i) \in V_C$. According to the assumption that $\exists y \in V_C \cup V_{NC}, y \neq \bot$ such that $y \sqsubseteq C(o) \sqcap \neg C(\bar{o})$ for all $\bar{o} \in \bar{O}$ and $o$ in KB, there exists node $y$ in $\mathcal{G}$ which yields paths from $\neg C(\bar{o}_i)$ to $y$ and $C(o)$ to $y$ for any $\bar{o}_i \in \bar{O}$. This implies that $Node_{\bot C(o)} \cap Node_{\bot \neg C(\bar{o}_i)} \neq \emptyset$ which corresponds to nonempty $\Omega_{m,P}$ of $C(\bar{o}_i)$ for any $\bar{o}_i \in \bar{O}$ and any path $P$ for an $m$. Hence, $T_{C(\bar{o}_i)}$ is a set in $SC$ for every $\bar{o}_i \in \bar{O}$. This implies that $SC$ is feasible.

Let $T_{\alpha_1}, T_{\alpha_2}, \ldots, T_{\alpha_k}$ be a feasible solution to $SC$. We know that $\bigcup_i T_{\alpha_i} = U$ and $\alpha_i \in sup_{C(o)}$. For each $\bar{o}_i$, there exists $j(i)$ such that $\bar{o}_i \in T_{\alpha_{j(i)}}$. Let $\alpha = \{\alpha_{j(i)}\}_{i=1}^n$. We next argue that $\alpha$ is an object evidence of falseness. The definition implies that $\alpha_{j(i)} \in P$ where $P \in Path_{C(\bar{o}_i)\top}$ which implies that $C(\bar{o}_i) \sqsubseteq \alpha_{j(i)}$. The first requirement in Definition 1 is therefore satisfied.

We next show the second property in Definition 1. Each $\alpha_{j(i)} \in sup_{C(o)}$ which is added in step 9 has to satisfy conditions in steps 6 and 8 according to Algorithm 1. For any path $P \in Path_{C(\bar{o}_i)^\top}$, $\Omega_{\alpha_{j(i)},P} \neq \emptyset$ implies that $\exists y \in V_C \cup V_{NC}$, $y \neq \bot$ such that there is a path from $C(o)$ to $y$ and a path from $\neg \alpha_{j(i)}$ to $y$. Hence, $\exists y \in V_C \cup V_{NC}$, $y \neq \bot$ such that $y \sqsubseteq C(o)$ and $y \sqsubseteq \neg \alpha_{j(i)}$ which is equivalent to $C(o) \not\sqsubseteq \alpha_{j(i)}$. As $C(o) \not\sqsubseteq \alpha_{j(i)}$ is assured for all $j(i)$, the second requirement in Definition 1 is satisfied. It is clear by construction that $k \geq |\alpha|$.

**Proof of Proposition 2:** Let $\alpha = \{\alpha_1, \alpha_2, ..., \alpha_k\}$ be feasible to $EP$.

By requirement 1 of $EP$, for every $\bar{o}_i \in \bar{O}$ there exists $\alpha_i$ such that $C(\bar{o}_i) \sqsubseteq \alpha_i$. From TBox classification which is used to construct the subsumption tree, we know that $a \sqsubseteq b$ if and only if there is a path from $b$ to $a$ for $a, b \in V_C$. Hence, a path $P \in Path_{C(\bar{o}_i)^\top}$ is one of the paths in step 4. The requirement 2 of $EP$ corresponds to $\Omega'$ computed and substituted to step 7 in Algorithm 1. If $\Omega'_{m,P} \neq \emptyset$ which is verified (by checking satisfiability of the concept $C(o) \sqcap \neg P_m$) in Step 8, $P_m$ considered as evidence $\alpha_i$ is added to $sup_{C(\bar{o}_i)}$ in step 9. By step 12 in Algorithm 1, $\alpha_i \in sup_{C(\bar{o}_i)}$ and thus $\alpha_i \in sup_{C(o)}$. Hence, there exists $T_{\alpha_i}$ for each $\alpha_i \in sup_{C(o)}$. Since $\alpha_i \in P$ where $P \in Path_{C(\bar{o}_i)^\top}$, $C(\bar{o}_i) \in T_{\alpha_i}$. We have $\bigcup_i T_{\alpha_i} \supseteq \bigcup_i C(\bar{o}_i) = \bar{O} = U$. This implies that $\alpha$ is feasible to $SC$. Since a feasible solution to $EP$ yields a feasible solution to $SC$, it implies that $Z_{SC} \leq Z_{EP}$. As $Z_{SC} \leq Z_{EP}$ and $Z_{SC} \geq Z_{EP}$ (from Proposition 1), $Z_{SC} = Z_{EP}$ is directly followed.

**Appendix D**

We manually observe extracted triplets in KG and compare them with their original statements. Based on our manual observation, incorrectly extracted triplets are provided in Table D1.

Table D1: Examples of incorrectly extracted triplets in KG based on SemRep

| **Original statements** | **Extracted triplets** |
|---|---|
| Six additional imino derivatives of pyridoxal have been studied, but none of these new compounds was as effective as PIH. | (Pyridoxal ; same as ; Prolactin Release-Inhibiting Hormone \| PEE1) |
| There was no significant different in the levels of G6PD activity in subjects with GdA or GdB. | (Glucose-6-phosphate dehydrogenase measurement, quantitative ; USES ; GDA) |
| Two methods for the removal of erythrocytes from buffy coats for the production of human leukocyte interferon. | (Erythrocytes ; PRODUCES ; human leukocyte interferon) |

**Appendix E**

According to the experiment, we observe matches among KG, Ollie and LSTM-ER models. Table E1 illustrates the number of matches among the different models.

Table E1: Number of matches among KG, Ollie and LSTM-ER models based on 10,000 triplets

| Number of triplets | KG | Ollie | LSTM-ER |
|---|---|---|---|
| 160 | × | × | × |
| 160 |   | × | × |
| 525 | × | × |   |
| 1500 | × |   | × |

**Appendix F**

We first identify the truthfulness of a lay triplet based on the PRA model for each relation type. A proportion of false triplets which is equivalent to the number of false triplets divided by total number of triplets is computed. Object candidates are computed for false triplets only. The total number of triplets, the false triplet proportion and the average number of object candidates for each relation types are summarized in Table F1.

Table F1: Statistics of truthfulness and object candidates obtained from the PRA model

| Relation type | Number of triplets | False triplets proportion | $Avg\ |\bar{O}|$ |
|---|---:|---:|---:|
| LOCATION_OF | 378 | 0.15 | 12.19 |
| ISA | 319 | 0.07 | 10.43 |
| PREDISPOSES | 292 | 0.33 | 8.68 |
| TREATS | 207 | 0.26 | 10.80 |
| CAUSES | 179 | 0.39 | 11.50 |
| AFFECTS | 116 | 0.41 | 14.69 |
| COEXISTS_WITH | 114 | 0.30 | 17.29 |
| PREVENTS | 110 | 0.25 | 7.86 |
| PART_OF | 69 | 0.25 | 7.65 |
| INTERACTS_WITH | 46 | 0.30 | 3.79 |
| INHIBITS | 45 | 0.20 | 7.67 |
| ASSOCIATED_WITH | 37 | 0.22 | 8.63 |
| AUGMENTS | 35 | 0.46 | 11.75 |
| USES | 35 | 0.09 | 7.67 |
| PRODUCES | 30 | 0.17 | 8.20 |
| DIAGNOSES | 22 | 0.14 | 14.00 |
| DISRUPTS | 21 | 0.48 | 9.70 |
| PRECEDES | 16 | 0.25 | 17.25 |
| METHOD_OF | 13 | 0.38 | 19.2 |

**Appendix G**

We compare the cardinality of candidates, the cardinality of evidence, and all elements in the set corresponding to "Al" and "Ma" in Table G1.

Table G1: A complete comparison of elements in evidence sets "Al" and "Ma"

| False Triplets | Al | | | Ma | | |
|---|---|---|---|---|---|---|
| | $|\bar{O}|$ | $|\alpha|$ | $\alpha$ | $|\bar{O}|$ | $|\alpha|$ | $\alpha$ |
| **Heparin** TREATS **Fever** | 2 | 1 | 'hemic system symptom' | 3 | 1 | 'hemic system symptom' |
| **Amiodarone** TREATS **Hepatitis C** | 5 | 1 | 'disease of anatomical entity' | 9 | 1 | 'disease of anatomical entity' |
| **Stress management** TREATS **Mitral Valve Prolapse** | 2 | 1 | 'nervous system disease' | 2 | 1 | 'nervous system disease' |
| **Capoten** TREATS **Coughing** | 12 | 1 | 'Disease, Disorder or Finding' | 9 | 1 | 'Disease, Disorder or Finding' |
| **Losartan** TREATS **Varicose Ulcer** | 5 | 3 | 'insulin resistance', 'hypertrophy', 'disease' | 9 | 2 | 'ischemia', 'disease' |
| **Echocardiography** DIAGNOSES **Hyperlipidemia** | 4 | 1 | 'disease of anatomical entity' | 4 | 1 | 'disease of anatomical entity' |
| **Platelet Size** DIAGNOSES **Anemia** | 4 | 1 | 'Cardiovascular Diseases' | 5 | 2 | 'Blood Platelet Disorders', 'Cardiovascular Diseases' |
| **Esophageal pH Monitoring** DIAGNOSES **Malignant breast neoplasm** | 11 | 5 | 'Biological Process', 'Finding', 'Non-Neoplastic Disorder', 'Neoplasm by Morphology', 'Digestive System Disorder' | 8 | 3 | 'Finding', 'Non-Neoplastic Disorder', 'Neoplasm by Morphology' |
| **Cholesterol measurement test DIANOSES Malignant breast neoplasm** | 15 | 5 | 'Biological Process', 'Mouse Disorder by Site', 'Finding', 'Non-Neoplastic Disorder', 'Dependence' | 4 | 2 | 'Non-Neoplastic Disorder by Site', 'Finding' |
| **Electrocardiogram DIAGNOSES Muscle strain** | 1 | 1 | 'cardiac disorder AE' | 1 | 1 | 'cardiac disorder AE' |
| **Caffeine CAUSES Gout** | 2 | 1 | 'Cell Physiological Phenomena' | 8 | 3 | 'Phenomena and Processes Category', 'Behavior and Behavior Mechanisms', 'Signs and Symptoms, Digestive' |
| **hypercholesterolemia CAUSES Neuropathy** | 7 | 2 | 'Finding', 'Non-Neoplastic Disorder' | 6 | 2 | 'Finding', 'Non-Neoplastic Disorder' |
| **Leukemia CAUSES Gout** | 3 | 2 | 'genetic disease', 'neoplasm (disease)' | 2 | 2 | 'genetic disease', 'neoplasm (disease)' |
| **Harpin CAUSES Cardiomegaly** | 7 | 2 | 'Pathologic Processes', 'Phenomena and Processes Category' | 2 | 1 | 'Cell Death' |
| **Ascorbic Acid CAUSES Senile Plaques** | 2 | 2 | 'Atherosclerosis', 'Abnormality of the cerebral ventricles' | 4 | 2 | 'Abnormality of digestive system physiology', 'Abnormality of nervous system physiology' |